\documentclass[journal]{IEEEtran}

\usepackage{cite}
\usepackage[utf8]{inputenc} 
\usepackage[T1]{fontenc}    
\usepackage{url}            
\usepackage{booktabs}       
\usepackage{amsfonts}       
\usepackage{nicefrac}       
\usepackage{microtype}      

\usepackage{graphicx}
\usepackage{comment}
\usepackage{amsmath,amssymb} 
\usepackage{color}
\usepackage{algorithm} 
\usepackage{algpseudocode} 
\usepackage{amsmath,graphicx,setspace}
\usepackage{multirow, array, amssymb,booktabs,makecell,xspace,bm}
\usepackage{amssymb}
\usepackage{pifont}
\usepackage{amsfonts}
\usepackage{arydshln}
\usepackage{makecell}
\usepackage{breqn}
\usepackage{amsmath,bm}

\newcommand{\cmark}{\ding{51}}%
\newcommand{\xmark}{\ding{55}}

\begin{document}

\title{Language-guided Learning for Object Detection Tackling Multiple Variations in Aerial Images}

\author{Sungjune Park, Hyunjun Kim, Beomchan Park, and Yong Man Ro,~\IEEEmembership{Senior Member,~IEEE}%
\thanks{S. Park, H. Kim, B. Park, and Y. M. Ro are with Image and Video Systems Lab., School of Electrical Engineering, Korea Advanced Institute of Science and Technology (KAIST), 291 Daehak-ro, Yuseong-gu, Daejeon, 34141, Republic of Korea (e-mail: sungjune-p@kaist.ac.kr; kimhj709@kaist.ac.kr; bpark0810@kaist.ac.kr; ymro@kaist.ac.kr). \\
Corresponding author: Y. M. Ro. (ymro@kaist.ac.kr)}}

\maketitle

\begin{abstract}
Despite recent advancements in computer vision research, object detection in aerial images still suffers from several challenges. One primary challenge to be mitigated is the presence of multiple types of variation in aerial images, for example, illumination and viewpoint changes. These variations result in highly diverse image scenes and drastic alterations in object appearance, so that it becomes more complicated to localize objects from the whole image scene and recognize their categories. To address this problem, in this paper, we introduce a novel object detection framework in aerial images, named \textbf{\underline{LANG}}uage-guided \textbf{\underline{O}}bject detection (\textbf{LANGO}). Upon the proposed language-guided learning, the proposed framework is designed to alleviate the impacts from both scene and instance-level variations. First, we are motivated by the way humans understand the semantics of scenes while perceiving environmental factors in the scenes (\textit{e.g.,} weather). Therefore, we design a visual semantic reasoner that comprehends visual semantics of image scenes by interpreting conditions where the given images were captured. Second, we devise a training objective, named relation learning loss, to deal with instance-level variations, such as viewpoint angle and scale changes. This training objective aims to learn relations in language representations of object categories, with the help of the robust characteristics against such variations. Through extensive experiments, we demonstrate the effectiveness of the proposed method, and our method obtains noticeable detection performance improvements.
\end{abstract}

\begin{IEEEkeywords}
Object detection in aerial images, Scene-level variation, Instance-level variation, Language-guided learning
\end{IEEEkeywords}

\IEEEpeerreviewmaketitle

\section{Introduction} \label{intro}
\begin{figure*}[t]
    \begin{center}
    \centerline{\includegraphics[width=0.95\linewidth]{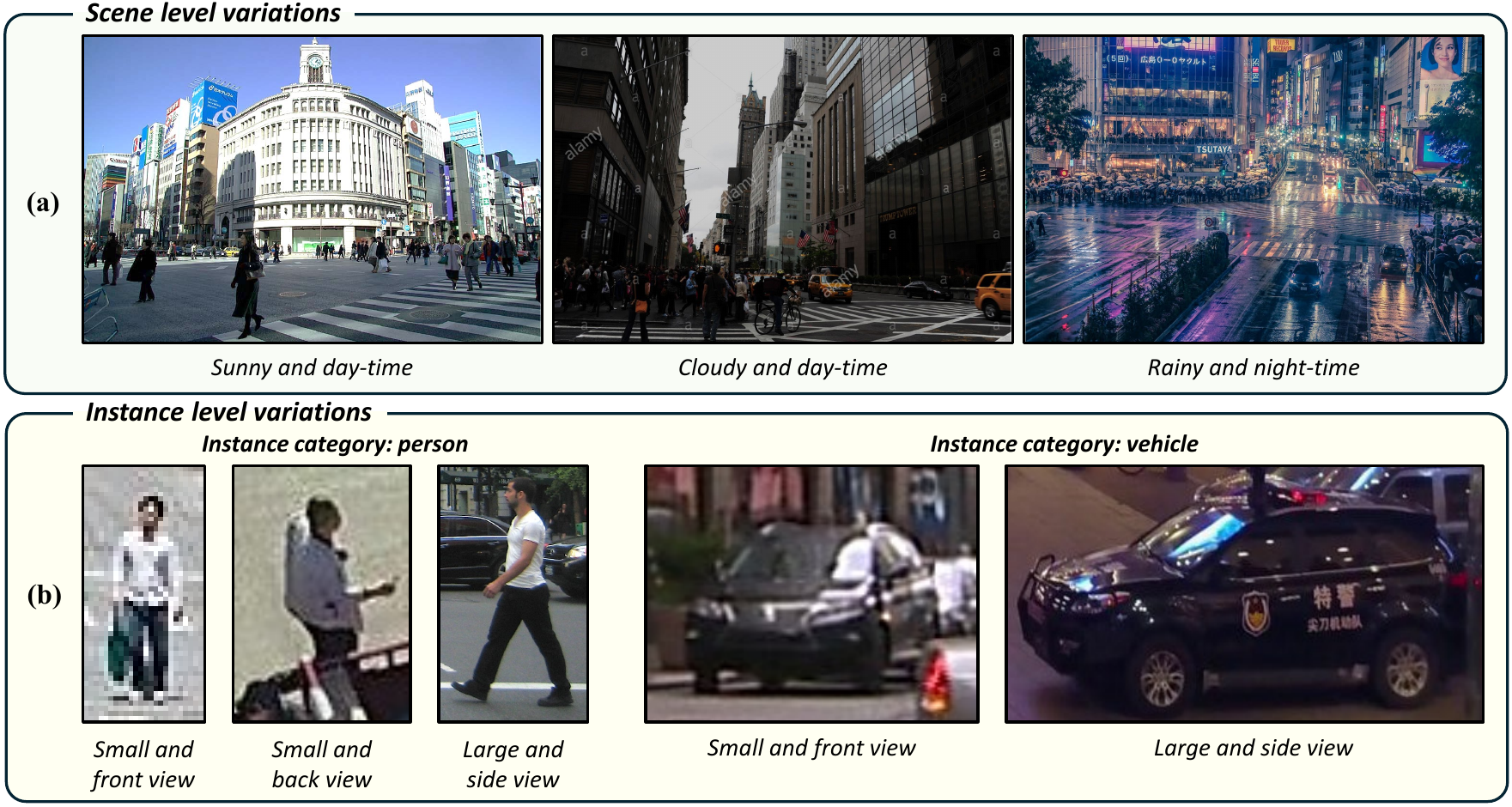}}
    \end{center}
    \caption{The examples illustrate that there exist multiple variations. (a) shows the scene-level variations (\textit{e.g.,} weather and illumination) and such environmental factors make the entire scenes varying. In the given examples, visual semantics differ from each other. The first image contains very clean weather and is taken at daytime, so that both object instances and backgrounds are well visible and distinguishable. On the other hand, since the third image is captured under rainy weather at nighttime, the scene is quite dim and objects are also less visible. Moreover, (b) describes that instance-level variations, such as changes in viewpoint angle and scale, make object instances look different even between objects within same categories. For example, \textit{person} instances are very different depending on their viewpoint angles and scales. Even though they are wearing white shirts and dark pants, their appearances varies. Also, two black vehicle also look different because of their scale and viewpoint.}
\label{fig1}
\end{figure*}

\IEEEPARstart{O}{bject} detection in aerial images has been considered one of the crucial research areas, while being adopted in a wide range of real-world applications, for example, safety surveillance and traffic management systems, and so on \cite{aerial-tcsvt, ufpmp, ceasc, aerial-tcsvt3, evorl, aerial-tcsvt2, aerial-tcsvt4}. However, despite the noticeable progress in deep learning and computer vision research, object detection in aerial images is still considered a challenging research task that suffers from several difficulties. One main problem stems from the inherent characteristics in aerial images that cause multiple types of variation and hinder robust object detection in aerial views. Such variations interrupt understanding entire images and perceiving object appearances. For example, environmental factors, such as weather and illumination conditions, drastically alter whole image scenes, so that it becomes difficult to comprehend visual semantics at the scene level. Moreover, viewpoint angles and altitude in aerial images severely change the visual appearances and scales of each object instance, and then it becomes more complicated to recognize objects. Fig.~\ref{fig1} illustrates how multiple types of variation affect visual semantics and appearances on both the image scene and object instance levels. In Fig.~\ref{fig1}(a), the image scenes change drastically depending on weather and illumination conditions (\textit{e.g.,} sunny, rainy, and night). Furthermore, Fig.~\ref{fig1}(b) shows how the angle of view and altitude alter the visual appearance of each object instance. Therefore, even though the object instances belong to the same object categories, their appearances vary significantly. For example, \textit{vehicle} instances look different according to their viewpoints and scales. While the left black vehicle is small (\textit{so blurred}) and faces toward showing its front headlight, the right black vehicle is large (\textit{so having a sharp outline}) and faces on side showing its windows.

However, rather than considering both scene- and instance-level variations simultaneously, current object detection studies in aerial images have focused on addressing one specific problem only \cite{ufpmp, dehazing, derain}. Fang \textit{et al.} \cite{dehazing} proposed object detection in a foggy weather condition with YOLO (ODFC-YOLO) trying to mitigate adverse impact of foggy weather. It consists of image dehazing and object detection networks together, so it utilizes a multi-task learning strategy to acquire clear image features even if foggy images are given. Similarly, Xi \textit{et al.} \cite{derain} presented CoDerainNet consisting of deraining and detection branches, so that it obtains rain-free image features. Furthermore, Huang \textit{et al.} \cite{ufpmp} introduced UFPMP-Det, a unified foreground packing object detection framework in aerial images. Since small objects are more easily cluttered from backgrounds, it aims to suppress the cluttering impact from backgrounds. Therefore, it is designed to find foreground candidate regions and concentrate on small objects effectively. Despite such efforts to mitigate variations in aerial images, these previous methods have handled variations in either scene level or instance level, overlooking the adverse impacts from both of them which are prevalent in aerial images.

Therefore, in this paper, we introduce a novel framework for object detection in aerial images, called \textbf{\underline{LANG}}uage-guided \textbf{\underline{O}}bject detection (\textbf{LANGO}), which incorporates language-guided learning to deal with multiple variations across scene and instance levels at the same time. The proposed language-guided learning approach exploits the strength of language for tackling each variation. When humans perceive and understand visual scenes, scene perception process starts from the global context that includes environmental factors, and then the process ends with recognizing objects properly even with their variations \cite{humanbrain}. Therefore, firstly, a visual semantic reasoner is designed to interpret the overall visual context of given aerial images. As human’s cognitive process does, it considers and understands environmental conditions, such as weather and illumination where images are captured. Therefore, it allows object detection framework to adapt to diverse variations in the scene level. Second, we devise a training objective, called relation learning loss, which is tailored to handle instance-level variations (\textit{e.g.,} changes in viewpoint and scale). Based on the observation that language representations describing each object are quite robust against such instance-level variations, the proposed training loss aims to learn their robust relations between object instances. Thus, it enables detection framework to recognize objects properly even with their alterations. We validate the effectiveness of the proposed method with extensive experiments on two widely used aerial object detection datasets, UAVDT and VisDrone, and the contributions of this paper are summarized threefold as follows:

\begin{itemize}
    \item We introduce a novel approach \textbf{\underline{LANGO}} for robust object detection in aerial images which utilizes language-guided learning to tackle scene and instance-level variations.
    \item We design 1) a visual semantic reasoner to comprehend varying semantics in the scene level and 2) a relation learning loss to learn the robust relations against appearance variations in the instance level.
    \item Extensive experiments and analyses corroborate the effectiveness of our method in challenging aerial object detection, showing considerable performance improvements.
\end{itemize}

\begin{figure*}[t]
    \begin{center}
    \centerline{\includegraphics[width=1.01\linewidth]{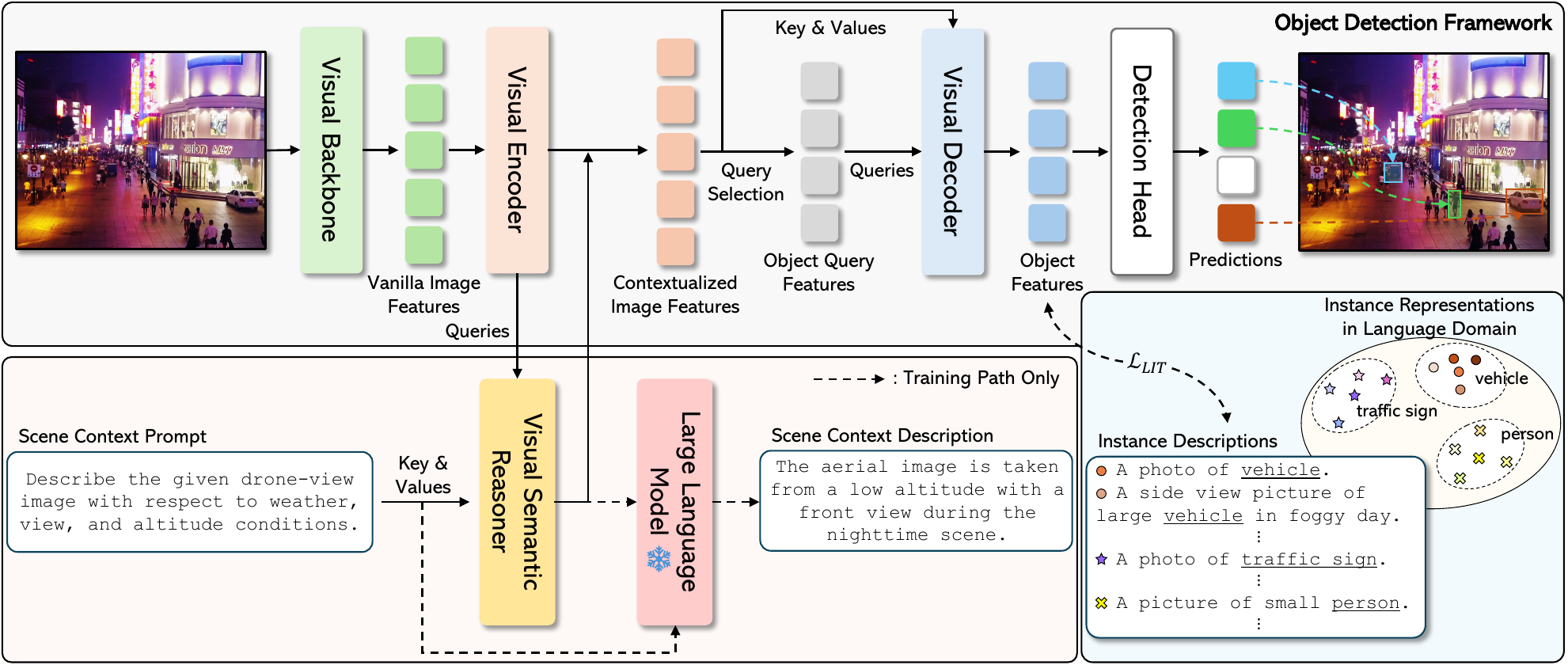}}
    \end{center}
    \caption{The overall architecture of the proposed method. While the object detection framework takes an input aerial image and extracts image features, we incorporate a visual semantic reasoner to comprehend visual semantics of the given image and adapt to diverse scene-level variations (\textit{e.g.,} weather and illumination). Moreover, when the detection framework is trained to predict object categories from visual object features, we additionally add a relation learning loss which is designed to resemble language representation relations between object instances, which are robust against instance-level variations.}
\label{fig2}
\end{figure*}

\section{Related Work}
\subsection{Object Detection in Aerial Images}
Object detection in aerial images refers to a task to localize and recognize objects in images taken from unmanned aerial vehicles. The primary problems to be solved are related with adverse impacts from environmental conditions and various object scales \cite{tph, ufpmp, dehazing, derain, aerial-tcsvt5}. Several previous methods have focused on alleviating such adverse impacts, especially from bad weather conditions. Such weather conditions include illumination changes like either daytime or nighttime, and those variations alter visual semantics of aerial images affecting the visibility of object instances. Fang \textit{et al.} \cite{dehazing} tried to mitigate the adverse effects of foggy weather by utilizing a multi-task learning method, so that a dehazing subnetwork is integrated into an object detection network. Therefore, it aims to minimize the feature gap between image features from foggy and clean weather conditions. Wu \textit{et al.} \cite{ppnet} proposed a plug-and-play aerial object detection network, that is PPNet. While a pretrained aerial object detection network is frozen, a hazy domain plug-in subnetwork is trained to extract foggy image features which are close with the features obtained from normal domain. Xi \textit{et al.} \cite{derain} designed a collaborative deraining network, called CoDerainNet, that adopts an additional deraining subnetwork to obtain rain-free image features. Furthermore, several methods have been proposed to deal with the scale variation problem of objects in aerial images. For example, Zhu \textit{et al.} \cite{tph} augmented a prediction head upon a transformer architecture, so that it utilizes a sparse local attention and enhances the feature quality of small objects. Huang \textit{et al.} \cite{ufpmp} presented UFPMP-Det, a unified foreground packing detection framework which consists of multi-proxy networks. To identify small objects efficiently, UFPMP-Det is designed to concentrate on the candidate regions at which small objects are likely to densely exist and suppress other less interesting regions. However, since such multiple types of variations interrupt robust object detection in aerial images, it is necessary to consider both types of variation at the same time. In this paper, we devise a language-guided learning for object detection in aerial images to tackle adverse impacts from both scene- and instance-level variations. To the best of our knowledge, it’s the first work to deal with such multiple variation problems in aerial object detection by exploiting the advantage

\subsection{Language-augmented Object Detection}
The proposed method is designed to employ the advantages of language, we introduce several existing works which utilize the power of language in object detection \cite{detclip, tcsvt-lang, da-pro, tcsvt-lang2}. In general, those studies adopt a generalization ability of language for open-vocabulary object detection. For example, Yao \textit{et al.} \cite{detclip} designed a concept dictionary consisting of prior concept knowledge and aligns textual embeddings with corresponding region embeddings. Jin \textit{et al.} \cite{llms-vlms} additionally adopted detailed descriptors for each category for text and region alignment. Lin \textit{et al.} \cite{generateu} introduced a generative and open-ended approach for object detection not to solely rely on alignment scores between regions and texts. Wei \textit{et al.} \cite{lenna} provided language instructions to identify specific objects and obtain candidate objects. Pi \textit{et al.} \cite{detgpt} also obtained candidate object category names by incorporating BLIP-2 \cite{blip2} and Vicuna \cite{vicuna}, and then Grounding-DINO \cite{grounding-dino} is utilized to localize objects. Moreover, several methods have adopted language information to address domain-specific problems in object detection. Vidit \textit{et al.} \cite{clipthegap} prepared illustrations of possible target domains, and then those descriptions are used for semantic feature augmentation. Li \textit{et al.} \cite{da-pro} placed learnable prompts added with category prompts for both source and target domains. Then their regions features are aligned with textual category prompt features. On the other hand, in this paper, we devise a language-guided learning to alleviate multiple variation problem, scene- and instance-level variations. It enables object detection framework to understand visual semantics of given aerial images and to obtain robust visual object features which resemble the robust relation of language object representations against variations.

\section{Proposed Method}
To deal with multiple variations at both scene and instance levels in aerial images, we introduce \textbf{\underline{LANG}}uage-guided \textbf{\underline{O}}bject detection framework, called \textbf{LANGO}. Fig.~\ref{fig2} illustrates the overall architecture of our proposed framework. This framework is built upon a transformer-based object detection network which is composed of visual backbone, visual encoder, visual decoder, and detection head. While taking input aerial images, the visual backbone network and visual encoder extract image features. While extracting image features, we consider that it would be beneficial to comprehend scene context of given aerial images. Therefore, we interleave a visual semantic reasoner with the visual encoder. The reasoner takes image features from the encoder and scene context prompt and is guided to interpret the visual semantics. By doing so, the reasoner generates contextualized image features, and the visual decoder utilizes these contextualized features for obtaining visual object features. After that, when the detection head predicts their categories, we guide the visual object features with the proposed relation learning loss $\mathcal{L}_{R}$. This relation learning loss uses diverse object instance descriptions. Since their representations are robust against instance-level variations, the visual object features are trained to learn and resemble their relations. Therefore, the visual object features can be also robust against the variations. The following subsections elaborate the details of the proposed visual semantic reasoner and relation learning loss, respectively.

\subsection{Visual Semantic Reasoner}
\begin{figure}[t]
    \begin{center}
    \centerline{\includegraphics[width=0.8\columnwidth]{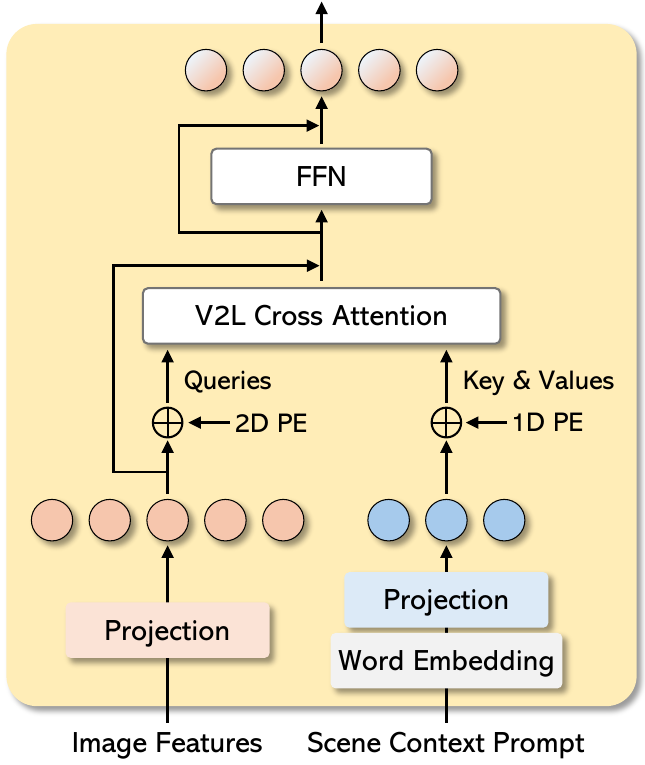}}
    \end{center}
    \caption{The details of the visual semantic reasoner consisting of vision-to-language (V2L) cross attention. It takes both image features and scene context prompt and utilizes them as query and key/value features, respectively.}
\label{fig3}
\end{figure}

The purpose of the visual semantic reasoner is to enable the visual encoder to capture the visual semantics and adapt to varying scene conditions properly, by reasoning the overall scene context of given aerial images. To this end, the visual semantic reasoner takes the image features from the visual encoder and a scene context prompt. As shown in Fig.~\ref{fig2}, the scene context prompt is the language prompt to ask the reasoner for the global scene context, and for example, it can be formulated as \textit{``Describe the given drone-view image with respect to weather, view, and altitude conditions.’’}. Fig.~\ref{fig3} shows the detailed architecture of the reasoner, and it is based on the vision-to-language (V2L) cross attention module taking the visual image features and language scene context prompt. Each of them is projected to align their feature space, and then 2D and 1D positional encoding (PE) are applied respectively. The image features are used as query features, and the scene context prompt features are used as key and value features. After passing through the reasoner, as shown in Fig.~\ref{fig2}, the output features are merged with the image features from the visual encoder via elementwise summation. At the same time, the output features are also fed into a pretrained large language model (LLM) along with a scene context prompt. This process aims to guide the reasoner to learn understanding visual semantics in the given scene context, and it is utilized during the training phase only. Then we guide LLM to generate a scene context description illustrating the visual semantics of the given image. For example, LLM generates \textit{``The drone-view image is taken from a low altitude with a front view during the nighttime scene.’’}. By doing so, the visual semantic reasoner learns to encode the scene contextual information from the image features. As mentioned above; by merging the output features from the reasoner with the image features from the encoder, the detection framework obtains the contextualized image features which capture diverse scene-level variations and understand the visual semantics from the given aerial scenes. Please note that LLM is employed during the training time only being frozen without any parameter updates, so that there is no additional cost from the usage of LLM during the inference time.

\begin{figure*}[t]
    \begin{center}
    \centerline{\includegraphics[width=0.95\linewidth]{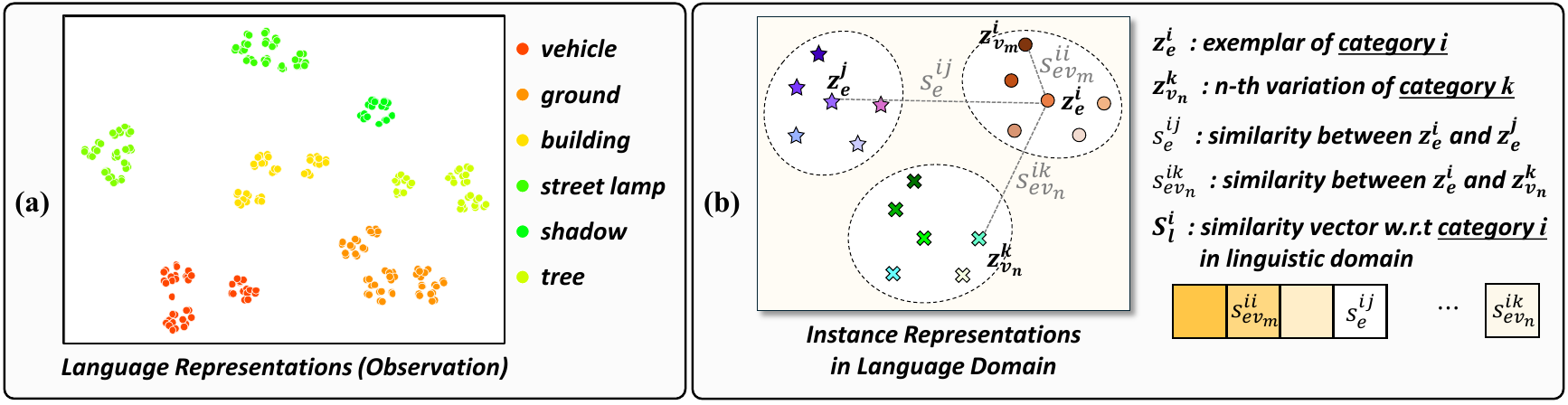}}
    \end{center}
    \caption{(a) shows t-SNE feature visualization results demonstrating that language instance representations are gathered together by object category, rather than being scattered by instance-level variation. These object categories include the categories from UAVDT dataset \cite{uavdt} and the others which probably exist in aerial images. (b) explains the process of deriving a similarity vector for the $i$-th category by using instance representations in language domain. These relation vector is used for the relation loss to guide the visual object features to learn the robust relations of language instance representations.}
\label{fig4}
\end{figure*}

\subsection{Relation Learning Loss}
The proposed relation learning loss is devised to guide the visual decoder to extract the visual object features which can be robust against objects’ appearance variations (\textit{i.e.,} instance-level variations). This training loss learns the object instance relations between their language representations, because those representations show robust characteristics. In other words, these representations are aggregated by object category and remain distinguishable from instances of other object categories, despite the presence of instance-level variations, such as changes in viewpoint angle and scale. To this end, we prepare the exemplar category language prompt, for example, \textit{``A photo of a vehicle.’’}. Then we augment this exemplar prompt with additional attributes, such as \textit{side view}, \textit{small}, \textit{foggy}, and so on, to obtain diverse instance descriptions illustrating their varying appearances. For example, \textit{``A side view photo of a small vehicle in a foggy day.’’} can be obtained. In this way, we curate the exemplar object category prompts and their variants in the language domain. We use the pre-trained sentence embedding model and extract their language representations, and then we observe how these representations are distributed. Fig.~\ref{fig4}(a) shows the t-SNE feature visualization of the exemplar and other variant representations for several object categories. 

After preparing such language representations including the exemplar and variant instance descriptions for each object category, we obtain a set of language representations $\boldsymbol{D^i}=\{\boldsymbol{d^i_{e}}, \boldsymbol{d^i_{v,n}}\}^{N}_{n=1}$, where $N$ is the number of representations for the $i$-th object category, and also each of $\boldsymbol{d^i_{e}}$ and $\boldsymbol{d^i_{v,n}}$ is the exemplar and variant representations, respectively, for the $i$-th object category. Even though these representations are already clustered together by their categories as shown in Fig.~\ref{fig4}(a), we place learnable categorical prompts $\boldsymbol{C} = \{\boldsymbol{c^i}\}^{N_C}_{i=1}$, where $N_C$ is the number of categories. These learnable categorical prompts are to make the language representations much more distinguishable between different category representations. The categorical prompts are merged with the language representations via elementwise summation as follows:
\begin{equation}
    \boldsymbol{z^i_{e}} = \boldsymbol{d^i_{e}} + \boldsymbol{c^i}, \quad
    \boldsymbol{z^i_{v,n}} = \boldsymbol{d^i_{v,n}} + \boldsymbol{c^i},
    \label{eq1}
\end{equation}
\noindent
where $\boldsymbol{z^i_{e}}$ and $\boldsymbol{z^i_{v,n}}$ are the exemplar and variant instance representations for the $i$-th category, respectively, and a set of the $i$-th category can be formulated $\boldsymbol{Z^i}=\{\boldsymbol{z^i_{e}}, \boldsymbol{z^i_{v,n}}\}^{N}_{n=1}$. After that, a set of instance representations across every object category is obtained, $\boldsymbol{Z} = \{\boldsymbol{Z^i}\}^{N_C}_{i=1}$. The learnable categorical prompts are randomly initialized and trained to make each element of $\boldsymbol{Z}$ to be gathered by category and repelled from the other categories through contrastive learning. Thus, we obtain $\boldsymbol{Z}$ which is more robust against instance-level variations. Note that we prepare $\boldsymbol{Z}$ before training the whole object detection framework.

To utilize the proposed relation learning loss, firstly, we acquire a similarity vector with respect to the $i$-th object category. Fig.~\ref{fig4}(b) describes how the similarity vector is obtained. For the $i$-th object category, we calculate cosine similarities between the exemplar representation $\boldsymbol{z^i_{e}}$ and every other representation from $\boldsymbol{Z}$. So that, we obtain a similarity vector $\boldsymbol{S^i_{l}}$ consisting of similarity values in language domain. For example, $s^{ii}_{ev_m}$ in the figure is the similarity between the exemplar instance representation of the $i$-th category ($\boldsymbol{z^i_{e}}$) and the $m$-th variant representation of the same $i$-th category ($\boldsymbol{z^i_{v_m}}$). Similarly, $s^{ij}_{e}$ is the similarity between the exemplar instance representations of the $i$-th and $j$-th categories ($\boldsymbol{z^i_{e}}$ and $\boldsymbol{z^j_{e}}$), and also $s^{ik}_{ev_n}$ is the similarity between the exemplar of the $i$-th category ($\boldsymbol{z^i_{e}}$) and the $n$-th variant of the $k$-th category ($\boldsymbol{z^k_{v_n}}$). We then normalize $\boldsymbol{S^i_{l}}$ via softmax and acquire a relation distribution for the $i$-th category $\boldsymbol{R^i_{l}}$. It contains the relations with the other representations including its variants and those of other categories in the language feature domain. And then, we also obtain the relations between visual object features and language representations. To this end, when we calculate the similarities between language representations, the visual object feature of the $i$-th category $\boldsymbol{o^i}$ replace the role of the exemplar representation $\boldsymbol{z^i_{e}}$. In other words, we calculate similarities between $\boldsymbol{o^i}$ and all other representations in language domain such as $\boldsymbol{z^i_{v_m}}$, $\boldsymbol{z^j_{e}}$, and $\boldsymbol{z^k_{v_n}}$ (except for the exemplar representation of $i$-th category $\boldsymbol{z^i_{e}}$). Thus, we obtain the similarity vector $\boldsymbol{S^i_{v}}$ using visual object features, in the same way described above. We also normalize it to obtain $\boldsymbol{R^i_{v}}$, a relation distribution between visual and language features with respect to the $i$-th category. Finally, we guide $\boldsymbol{R^i_{v}}$ to learn the relations of $\boldsymbol{R^i_{l}}$ by using Kullback-Leiber (KL) divergence as follows:
\begin{equation}
    \mathcal{L}_{R} = \frac{1}{N_C} \sum_{i=1}^{N_C} D_{KL}(\boldsymbol{R^i_{l}}||\boldsymbol{R^i_{v}}),
    \label{eq2}
\end{equation}
\noindent
where $\mathcal{L}_{R}$ is the relation learning loss averaged across object categories. Due to the proposed training loss, the visual object features can learn the relations from the language domain which are robust against diverse instance-level variations, so that it helps for the visual decoder to extract robust visual object features.

\subsection{Total Training Objective}
Finally, we train the entire proposed framework in an end-to-end manner by utilizing the conventional object detection losses for each classification and localization ($\mathcal{L}_{cls}$ and $\mathcal{L}_{loc}$), and the relation learning loss ($\mathcal{L}_{R}$). Therefore, the final training objective is as follows:
\begin{equation}
    \mathcal{L}_{final} = \mathcal{L}_{cls} + \mathcal{L}_{bbox} + \mathcal{L}_{R}.
    \label{eq3}
\end{equation}

\section{Experimental Setup}
\label{setup}
\subsection{Datasets}
\label{data}
To validate the effectiveness of our proposed method for tackling scene- and instance-level variations on object detection in aerial images, we use two public drone-view object detection datasets, UAVDT \cite{uavdt} and VisDrone \cite{visdrone}. Also, we elaborate how to prepare the scene context prompts and instance descriptions for each dataset in following subsections.

\subsubsection{UAVDT Dataset} UAVDT is one of the widely used drone-view object detection dataset, and it contains 24,143 train images and 16,592 test images. The images are captured by a drone from low to high altitude, and the average image resolution is about 1,080 $\times$ 540. It provides annotations of weather conditions, flying altitude, and camera views, among others. There are three categories: \textit{car}, \textit{truck}, and \textit{bus}. However, since there is a severe category imbalance problem inherent in the dataset, we merge these three categories into a single \textit{vehicle} category following previous methods \cite{uavdt, delving}. As rich annotations are provided regarding the environmental factors (\textit{i.e.,} weather, altitude, and camera view), we utilize these annotations to generate scene context description when guiding the visual semantic reasoner. As scene context prompts, we obtain four prompt options: 1) \textit{``Describe the given drone-view image with respect to weather, view, and altitude conditions.’’}, \textit{``What are the weather, view, and altitude conditions in the given drone-view image?’’}, \textit{``Illustrate the weather, view, and altitude conditions in the given drone-view image.’’}, \textit{``Describe the weather, view, and altitude conditions in which the given drone-view image is captured.’’}. We randomly sample one prompt at every iteration. As a scene context description, we use the form of \textit{``The drone-view image is taken from a [alt] altitude with a [view]’ view during the [weat] scene.’’}. \textit{[alt]} can be \textit{low}, \textit{medium}, and \textit{high}, \textit{[view]} can be \textit{side}, \textit{front}, and \textit{bird}, and \textit{[weat]} can be \textit{clean}, \textit{night}, and \textit{foggy}. For the evaluation metric, we adopt PASCAL VOC Average Precision (AP) metric with a 0.7 intersection of union (IoU) threshold to perform a fair comparison with previous methods \cite{uavdt, delving}.

\subsubsection{VisDrone Dataset} VisDrone is another large drone-view object detection dataset which is composed of diverse and challenging conditions, such as weather and illumination. It includes 6,471, 548, and 1,580 images for training, validation, and testing, respectively, and contains 10 object categories: \textit{pedestrian}, \textit{people}, \textit{bicycle}, \textit{car}, \textit{van}, \textit{truck}, \textit{tricycle}, \textit{awning-tricycle}, \textit{bus}, and \textit{motor}. Since there is no annotation with respect to the attribute variations, we adopt Meteor \cite{meteor} to obtain pseudo labels for the time the images are taken. Therefore, we obtain pseudo \textit{daytime} and \textit{nighttime} labels which are paired with each image. As scene context prompts, we use four prompt options: 1) \textit{``Describe the given drone-view image with respect to time that the image is captured.’’}, \textit{``What is time condition for the given drone-view image?’’}, \textit{``Illustrate the time which the given drone-view image is taken.’’}, \textit{``Describe which time the given drone-view image is captured.’’}. We randomly sample one prompt at every iteration. As a scene context description, we use the form of \textit{``The drone-view image is taken at [time] time.’’}. \textit{[time]} is either \textit{day} or \textit{night}. Following previous works \cite{delving}, we use the validation set for fair evaluation and measure AP$_{50}$ performance based on MS-COCO protocol \cite{coco}, and we denote it as AP for a brevity.

\subsection{Implementation Detail}
\label{impl}
We implement the proposed method using PyTorch library \cite{pytorch}, and our object detection framework is built upon the transformer-based object detection architecture, RT-DETR \cite{rtdetr}. We conduct experiments with ResNet-50 (R50) and ResNet-101 (R101) visual backbones, respectively. We also use DDQ-DETR \cite{ddq} with R50 and Swin-L visual backbones for the detection framework extension experiment. Even if not stated, we use RT-DETR as our baseline object detection framework. When guiding the visual semantic reasoner, we adopt Mistral-7B-Intsruct-v0.3 \cite{mistral-7b, mist-7b-inst-v0.3} as large language model (LLM). The output features of the visual semantic reasoner are passed through 5$\times$5 convolution layers to reduce the sequence length of visual image features and to be adapted to the input domain of LLM. Additionally, we curate instance descriptions based on \textit{``A [view] view picture/photo of a [scale] [category] on a [wea] scene.’’}. Here, \textit{[view]}, \textit{[scale]}, \textit{[category]}, and \textit{[wea]} are the place holders for the specific conditions of viewpoint, scale, object category, and weather, respectively. For the category, we utilize predefined object categories from both datasets and other common object categories which can be usually observed in aerial images, such as \textit{ground}, \textit{building} \textit{tree}, and \textit{traffic sign}. When obtaining language representations for instance exemplar and variant descriptions, we adopt pretrained sentence embedding model, called \textit{all-mpnet-base-v2} \cite{mpnet} which is trained to generate and aggregate output word token embeddings to represent an entire sentence. The visual decoder consists of 6 transformer layers to refine its outputs, and the proposed relation loss is applied to every layer. We train the whole detection framework in an end-to-end manner on 4 A100 GPU devices along with 4 image batches per device by using AdamW optimizer. The optimizer has $4e-4$ learning rate, and there is 2,000 warm-up iterations. The total number of training epochs are 14 for UAVDT and 108 for VisDrone, and we also augment data using multi-scaling, photometric distortion, and horizontal flip techniques.

\renewcommand{\arraystretch}{1.1}
\begin{table}[t]
    \centering
    \caption{Comparison of our method with state-of-the-art methods on UAVDT test set.}
    \renewcommand{\tabcolsep}{3.0mm}
    \resizebox{0.99\linewidth}{!}
    {\small
        \begin{tabular}{lcc}
            \toprule[1.1pt]
            \bf Methods & \bf Backbone & \bf AP  \\
            \midrule
            CADNet (TCSVT 2019) \cite{cadnet} & VGG16 & 43.7 \\
            GANet (ACMMM 2020) \cite{ganet} & VGG16 & 46.8 \\
            OSR-FPN (TCSVT 2022) \cite{osr-fpn} & MobileNet-V2 & 47.3 \\
            NDFT (ICCV 2019) \cite{delving} & R101 & 47.9 \\
            A-NDFT (CVPR 2021) \cite{a-ndft} & R101 & 48.1 \\
            SpotNet (CRV 2020) \cite{spotnet} & Hourglass-104 & 52.8 \\
            FLSL (NIPS 2023) \cite{flsl} & ViT-S/16 & 53.1 \\
            FLSL (NIPS 2023) \cite{flsl} & ViT-B/16 & 53.5 \\
            Focus\&Detect (SPIC 2022) \cite{focus} & X101 & 54.2 \\
            FLSL (NIPS 2023) \cite{flsl} & ViT-S/8 & 55.2 \\
            \midrule
            \bf LANGO (Ours) & \bf R50 & \bf 57.9 \\
            \bf LANGO (Ours) & \bf R101 & \bf 58.6 \\
            \bottomrule[1.1pt]
        \end{tabular}
    }
    \label{tab1}
\end{table}

\begin{table}[t]
    \centering
    \caption{Comparison of our method with state-of-the-art methods on VisDrone validation set.}
    \renewcommand{\tabcolsep}{4.6mm}
    \resizebox{0.99\linewidth}{!}
    {\small
        \begin{tabular}{lcccc}
            \toprule[1.1pt]
            \bf Methods & \bf Backbone & \bf AP \\
            \midrule
            CEASC (CVPR 2023) \cite{ceasc} & R18 & 50.7 \\
            DSHNet (WACV 2021) \cite{dshnet} & R50 & 51.8 \\
            NDFT (ICCV 2019) \cite{delving} & R101 & 52.8 \\
            FLDet-S (TCSVT 2024) \cite{fldet} & CSPNet & 54.8 \\
            YOLC (TITS 2024) \cite{yolc} & R50 & 55.0 \\
            OGMN (P\&RS 2023) \cite{ogmn} & R101 & 57.7 \\
            OGMN (P\&RS 2023) \cite{ogmn} & X101 & 59.7 \\
            PRDet (TCSVT 2022) \cite{prdet} & X101 & 60.8 \\
            UFPMP (AAAI 2022) \cite{ufpmp} & R50 & 62.4 \\
            UFPMP (AAAI 2022) \cite{ufpmp} & R101 & 63.2 \\
            YOLC (TITS 2024) \cite{yolc} & HRNet & 63.7 \\
            \midrule
            \bf LANGO (Ours) & \bf R50 & \bf 64.4 \\
            \bf LANGO (Ours) & \bf R101 & \bf 64.8 \\
            \bottomrule[1.1pt]
        \end{tabular}
    }
    \label{tab2}
\end{table}

\section{Experimental Result}
\label{expr}
\subsection{Comparison with State-of-the-art Methods}
\label{comp}
We validate the effectiveness of the proposed approach by comparing the aerial object detection performance with existing state-of-the-art methods on UAVDT and VisDrone datasets. For the comparison on UAVDT, we compare our method with CADNet \cite{cadnet}, NDFT \cite{delving}, GANet \cite{ganet}, SpotNet \cite{spotnet}, A-NDFT \cite{a-ndft}, Focus\&Detect \cite{focus}, and FLSL \cite{flsl}. The evaluation is conducted on the test dataset of UAVDT, and TABLE~\ref{tab1} shows the comparison results. As shown in the table, the proposed method outperforms the previous methods, achieving state-of-the-art performance. With ResNet-50 (R50) and ResNet-101 (R101) visual backbones, our framework obtains 57.9 AP and 58.6 AP, respectively, which are 2.7 AP and 3.4 AP higher than the previous state-of-the-art method \cite{flsl}. Furthermore, we also evaluate the detection performance on the validation set of VisDrone dataset, and TABLE~\ref{tab2} shows the comparison results with the existing methods: NDFT \cite{delving}, DSHNet \cite{dshnet}, UFPMP \cite{ufpmp}, OGMN \cite{ogmn}, CEASC \cite{ceasc}, and YOLC \cite{yolc}. The table shows that our method achieves notable performances. With each R50 and R101 visual backbones, our method obtains 64.4 AP and 64.8 AP. Such comparative experimental results quantitatively corroborate the effectiveness of our language-guided learning approach which tackles multiple scene- and instance-level variations. Also, please note that we do not compare our method with the existing methods which adopts crop\&zoom technique (which enlarge some partial regions) and multi-scale testing to boost their performances.

\subsection{Ablation Study}
\label{abl}
We also conduct ablation study to verify the effectiveness of each proposed component: the visual semantic reasoner and the relation learning loss. This ablation study is performed on UAVDT test set and VisDrone validation set, and also we use R50 visual backbone. TABLE~\ref{tab3} shows the ablation study results. From the baseline detection framework where the proposed method is not applied at all, each of the components shows remarkable performance improvements, respectively. Those components obtain 2.6 AP and 2.4 AP improvements on UAVDT, and also 3.8 AP and 3.3 AP improvements on VisDrone. Finally, when those components are adopted together, the detection framework shows the best performance, 57.9 AP and 64.4 AP, on UAVDT and VisDrone.

\begin{table}[t]
    \centering
    \caption{Ablation study on UAVDT and VisDrone to verify the effectiveness of each proposed component, the visual semantic reasoner and the relation learning loss $\mathcal{L}_{R}$.}
    \renewcommand{\tabcolsep}{6.0mm}
    \resizebox{0.99\linewidth}{!}
    {\small
        \begin{tabular}{cccc}
        \toprule[1.1pt]
        \multirow{2}{*}{\bf Reasoner} & \multirow{2}{*}{$\boldsymbol{\mathcal{L}_{R}}$} & \bf UAVDT & \bf VisDrone \\
          &  & \bf AP & \bf AP \\
        \midrule
        - & - & 52.8 & 58.8 \\ \hdashline
        \rule{0pt}{10pt}
        \cmark & - & 55.4 & 62.6 \\
        - & \cmark & 55.2 & 62.1 \\
        \cmark & \cmark & \bf 57.9 & \bf 64.4 \\
        \bottomrule[1.1pt]
        \end{tabular}
    }
    \label{tab3}
\end{table}

\begin{figure}[t]
    \begin{center}
    \centerline{\includegraphics[width=\columnwidth]{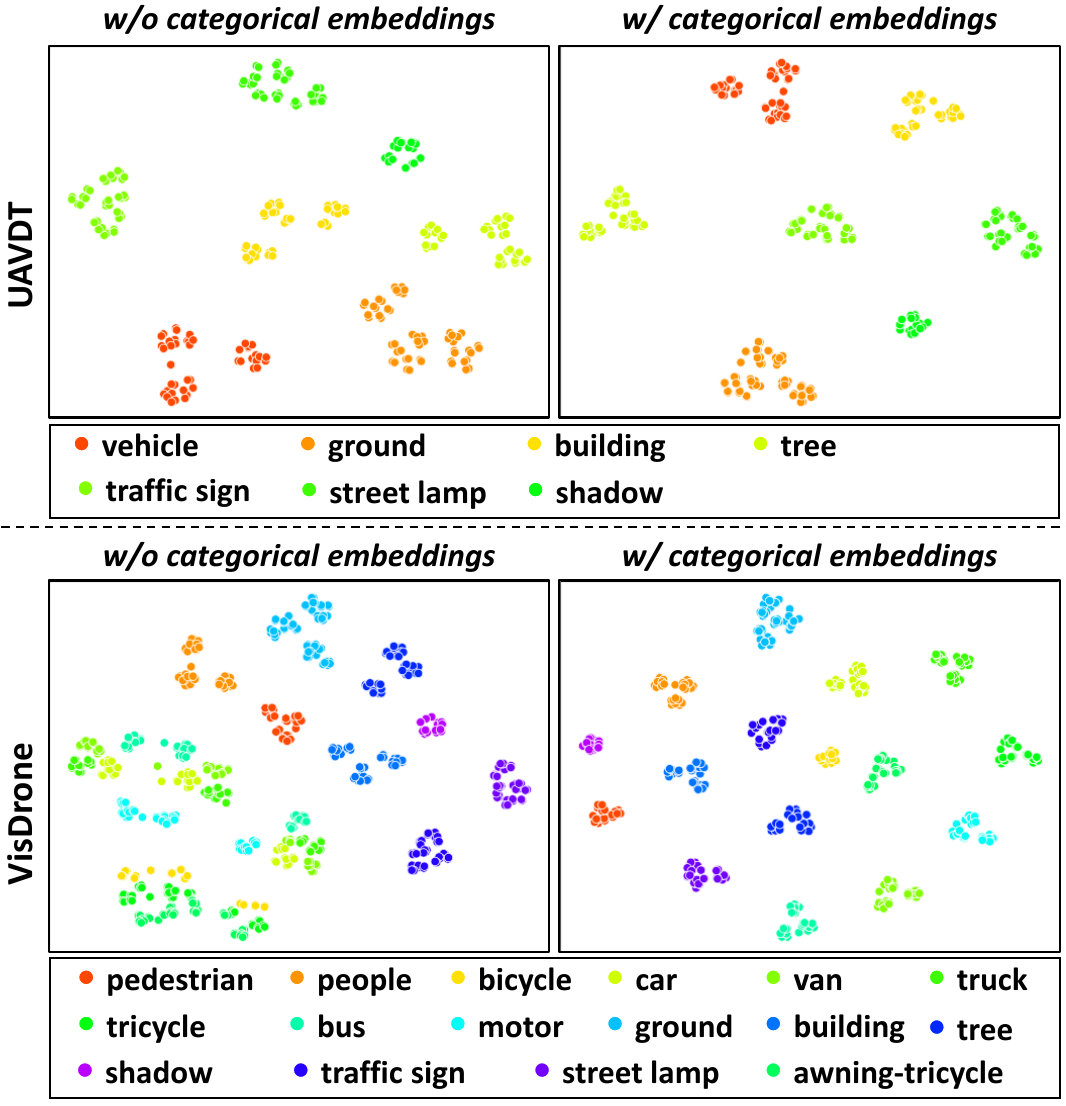}}
    \end{center}
    \caption{The t-SNE visualization of language instance representations for object categories in UAVDT and VisDrone. Based on the language representations which are distinguishable from each other category, the learnable categorical prompts help them more distinct and robust against instance-level variations.}
\label{fig5}
\end{figure}

\begin{figure*}[t]
    \begin{center}
    \centerline{\includegraphics[width=1.01\linewidth]{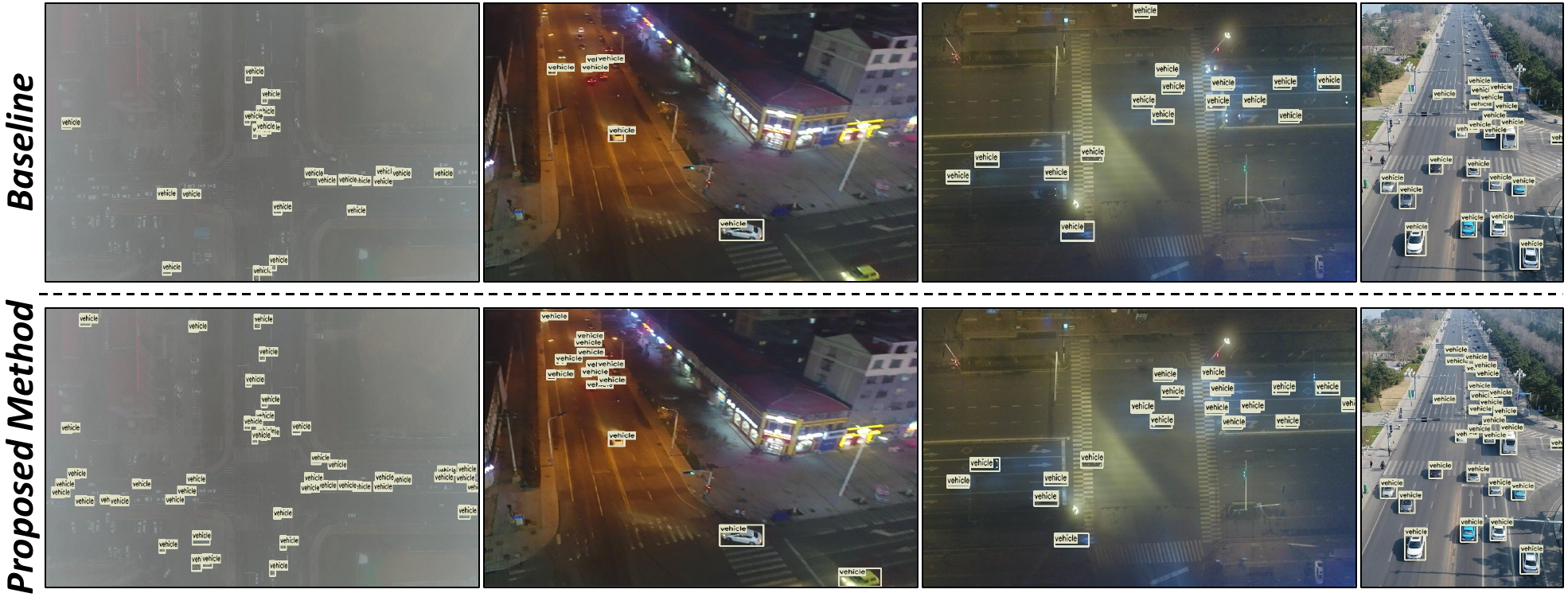}}
    \end{center}
    \vspace{-0.3cm}
    \caption{The qualitative visualization results on severe environmental conditions from UAVDT test set. The upper and lower figures show the detection results from the baseline and the proposed framework, respectively. The figures are cropped and enlarged from the original images for the better visualization. As shown in the figures, our proposed method performs more robust object detection in aerial images.}
\label{fig6}
\end{figure*}

\begin{table*}[h]
    \centering
    \caption{Detection performance analysis on UAVDT test set. Following \cite{ganet}, we measure the detection performances on nine variations subsets separately. Both R50 and R101 backbones are used, and our frameworks show robust performances across every variation.}
    \renewcommand{\tabcolsep}{2.5mm}
    \resizebox{0.95\linewidth}{!}
    {\small
        \begin{tabular}{lccccccccccc}
            \toprule[1.1pt]
            \bf Methods & \bf AP & \bf AP$_{day}$ & \bf AP$_{night}$ & \bf AP$_{fog}$ &
            \bf AP$_{low}$ & \bf AP$_{med}$ & \bf AP$_{high}$ & \bf AP$_{front}$ &
            \bf AP$_{side}$ & \bf AP$_{bird}$ \\
            \midrule
            GANet \cite{ganet} & 46.8 & 52.4 & 58.4 & 30.8 & 58.2 & 52.1 & 26.2 & 48.5 & 54.4 & 36.0 \\
            \midrule
            \bf LANGO (R50) & 57.9 & 64.7 & 75.5 & 37.5 & 72.7 & 65.8 & 30.5 & 58.1 & 68.1 & 44.0 \\
            \bf LANGO (R101) & 58.6 & 65.8 & 75.6 & 37.2 & 71.9 & 67.3 & 31.2 & 58.3 & 69.7 & 43.9 \\
            \bottomrule[1.1pt]
        \end{tabular}
    }
    \label{tab4}
\end{table*}

\subsection{Language Instance Representation Analysis}
\label{lang.ins.repr}
To mitigate instance-level variations, we devise the relation learning loss to learn the relations between instance representations which are extracted from language instance descriptions. This loss is motivated by that those representations are robust against objects’ appearance variations. Fig.~\ref{fig5} shows the feature distributions on UAVDT (top) and VisDrone (bottom) for both scenarios: the learnable categorical prompts are applied (\textit{right}) or not (\textit{left}). As shown in the left figures, even though the learnable categorical prompts are not adopted, the language instance representations are properly separated from other categories. Therefore, language representations can be good candidates to guide visual object features to be robust against variations. Furthermore, we explore how the representations change when the learnable categorical prompts are employed. As shown in the right figures on each dataset, the instance representations become much more distinguishable, so that they are well aggregated by category. Thus, finally, we use these instance representations for the proposed relation loss. Please note that the categories include other possible background categories, such as streetlamp and traffic sign, helping the visual object features distinguish the target categories (\textit{e.g.,} vehicle, pedestrian, \textit{etc.}) from backgrounds effectively.

\subsection{Visualization Result}
\label{vis.res}
Fig.~\ref{fig6} illustrates the visualization examples of detection results from UAVDT test set. We compare the baseline and the proposed method using R101 visual backbone network. As shown in the first scene, the image is under a very severe foggy day, so that the baseline could not capture object instances properly. On the other hand, the proposed method shows the improved detection results detecting more objects successfully. Moreover, even in the nighttime, our framework shows more robust detection performance qualitatively.

\subsection{Performance Analysis across Diverse Variations}
\label{perf.anal.vari}
To validate the effectiveness of the proposed method across diverse variations, we analyze the performance for each variation, separately. For this analysis, we use the test set of UAVDT. Following the previous method \cite{ganet}, we measure detection performance across nine individual variation subsets with respect to weather, altitude, and viewpoint angle. Specifically, these variations include \textit{daytime}', \textit{nighttime}', \textit{foggy}', \textit{low-altitude}', \textit{medium-altitude}', \textit{high-altitude}', \textit{front-view}', \textit{side-view}', and `\textit{bird-view}' conditions. The weather variation mainly causes illumination changes, the altitude affects scale changes of object instances, and the viewpoint could alter both scene- and instance-level changes. TABLE~\ref{tab4} shows the performance analysis results across these 9 variation subsets, individually. As shown in the table, our method achieves noticeable performances across all variation subsets due to the proposed language-guided learning.

\section{Discussion}
\label{disc}
\subsection{Detection Framework Extension}
\label{frame.ext}
While we primarily adopt RT-DETR \cite{rtdetr} with R50 and R101 as our baseline object detection framework, we also extend another detection framework, DDQ-DETR \cite{ddq} with R50 and Swin-L visual backbones, to discuss the applicability of our method to other frameworks and backbone networks. The evaluation is conducted on UAVDT test set. TABLE~\ref{tab5} shows the extension results, and we also compare the performance with several state-of-the-art methods, Focus\&Detect \cite{focus} and FLSL \cite{flsl}. As shown in the table, the proposed method shows noticeable performance improvements, from 48.2 AP to 56.0 AP with R50 backbone networks, and from 48.6 AP to 57.2 AP with Swin-L backbone networks, respectively.

\begin{table}[t]
    \centering
    \caption{The object detection framework extension results on UAVDT test set. DDQ-DETR with R50 and Swin-L visual backbone networks are adopted.}
    \renewcommand{\tabcolsep}{2.3mm}
    \resizebox{\linewidth}{!}
    {\small
        \begin{tabular}{lccc}
            \toprule[1.1pt]
            \bf Methods & \bf Backbone & \bf Proposed Method & \bf AP  \\
            \midrule
            FLSL \cite{flsl} & ViT-S/16 & -- & 53.1 \\
            FLSL \cite{flsl} & ViT-B/16 & -- & 53.5 \\
            Focus\&Detect \cite{focus} & X101 & -- & 54.2 \\
            FLSL \cite{flsl} & ViT-S/8 & -- & 55.2 \\
            \midrule
            DDQ-DETR & R50 & \xmark & 48.2 \\
            DDQ-DETR & R50 & \cmark & \bf 56.0 \\
            \hdashline
            DDQ-DETR & Swin-L & \xmark & 48.6 \\
            DDQ-DETR & Swin-L & \cmark & \bf 57.2 \\
            \bottomrule[1.1pt]
        \end{tabular}
    }
    \label{tab5}
\end{table}

\begin{table}[t]
    \centering
    \caption{The comparison of inference speed. We measure the inference speed on NVIDIA RTX 3090 GPU device, and AP is measured on VisDrone validation set.}
    \renewcommand{\tabcolsep}{5.5mm}
    \resizebox{\linewidth}{!}
    {\small
        \begin{tabular}{lcc}
            \toprule[1.1pt]
            \bf Methods & \bf AP & \bf Inference Speed (s/img) \\
            \midrule
            UFPMP \cite{ufpmp} & 62.4 & 0.165 \\
            PRDet \cite{prdet} & 53.9 & 0.195 \\
            YOLC \cite{yolc} & 63.7 & 0.346 \\
            \midrule
            Baseline (R50) & 58.8 & 0.050 \\
            \bf LANGO (Ours) & \bf 64.4 & \bf 0.057 \\
            \bottomrule[1.1pt]
        \end{tabular}
    }
    \label{tab6}
\end{table}

\subsection{Inference Speed}
\label{inf.speed}
We further discuss inference speed, that is, the time taken for processing one aerial image. We measure the inference speed on NVIDIA RTX 3090 GPU device and compare with several existing methods, UFPMP \cite{ufpmp}, PRDet \cite{prdet}, and YOLC \cite{yolc}. Moreover, we also measure the inference speed of our baseline, that is, RT-DETR with R50. TABLE~\ref{tab6} shows the measured inference speed, and \textbf{AP} denotes the performance on VisDrone validation set. Built upon the noticeably efficient RT-DETR \cite{rtdetr}, our proposed method only requires additional 0.007 seconds to process one aerial image, showing superior efficiency than the previous methods.

\subsection{Social Impact}
\label{soc.imp}
In this section, we discuss several social impacts caused by aerial object detection research. In particular, it can be adopted in various real-world public systems for enhancing public safety, traffic monitoring, environmental management (\textit{e.g.,} disaster monitoring), and so on. It enables a rapid assessment by providing crucial information in a wide range of scenes. However, there might be also privacy concerns which leads to unauthorized collection and misuse. Therefore, it is necessary to establish very strict ethical guidelines before deploying such a technology. This awareness ensures a balance between positive social benefits from aerial object detection and individual rights protection.

\section{Conclusion}
In this paper, we proposed a novel object detection framework in aerial images, called LANGO, by utilizing the language-guided learning approach. The proposed method is motivated by the way humans perceive visual scenes. Like the perception process of humans, our method is designed to understand environmental factors, such as weather, in the global scene. So that, the visual semantic reasoner is integrated to contextualize visual semantics across entire images by reasoning which conditions the given images are captured. Moreover, like humans can consider how objects look like and perceive which category objects belong to despite their varying appearances, the relation loss is also devised to learn the robust relations between language instance representations which are robust against instance-level variations. So that, the visual object features also become robust against objects’ varying appearances. Through extensive experiments, the proposed method shows the effectiveness on two public aerial object detection datasets, UAVDT and VisDrone, outperforming other previous methods. Consequently, we hope that our method can provide useful insights to a variety of research where diverse types of variations exist and many other vision tasks where language can deliver beneficial information.

\ifCLASSOPTIONcaptionsoff
  \newpage
\fi

\bibliographystyle{IEEEtran}
\bibliography{IEEEabrv,main}

\begin{thebibliography}{10}
\providecommand{\url}[1]{#1}
\csname url@samestyle\endcsname
\providecommand{\newblock}{\relax}
\providecommand{\bibinfo}[2]{#2}
\providecommand{\BIBentrySTDinterwordspacing}{\spaceskip=0pt\relax}
\providecommand{\BIBentryALTinterwordstretchfactor}{4}
\providecommand{\BIBentryALTinterwordspacing}{\spaceskip=\fontdimen2\font plus
\BIBentryALTinterwordstretchfactor\fontdimen3\font minus \fontdimen4\font\relax}
\providecommand{\BIBforeignlanguage}[2]{{%
\expandafter\ifx\csname l@#1\endcsname\relax
\typeout{** WARNING: IEEEtran.bst: No hyphenation pattern has been}%
\typeout{** loaded for the language `#1'. Using the pattern for}%
\typeout{** the default language instead.}%
\else
\language=\csname l@#1\endcsname
\fi
#2}}
\providecommand{\BIBdecl}{\relax}
\BIBdecl

\bibitem{aerial-tcsvt}
A.~ElTantawy and M.~S. Shehata, ``Krmaro: Aerial detection of small-size ground moving objects using kinematic regularization and matrix rank optimization,'' \emph{IEEE Transactions on Circuits and Systems for Video Technology}, vol.~29, no.~6, pp. 1672--1686, 2018.

\bibitem{ufpmp}
Y.~Huang, J.~Chen, and D.~Huang, ``Ufpmp-det: Toward accurate and efficient object detection on drone imagery,'' in \emph{Proceedings of the AAAI Conference on Artificial Intelligence}, vol.~36, no.~1, 2022, pp. 1026--1033.

\bibitem{ceasc}
B.~Du, Y.~Huang, J.~Chen, and D.~Huang, ``Adaptive sparse convolutional networks with global context enhancement for faster object detection on drone images,'' in \emph{Proceedings of the IEEE/CVF Conference on Computer Vision and Pattern Recognition}, 2023, pp. 13\,435--13\,444.

\bibitem{aerial-tcsvt3}
Y.~Yao, G.~Cheng, C.~Lang, X.~Yuan, X.~Xie, and J.~Han, ``Hierarchical mask prompting and robust integrated regression for oriented object detection,'' \emph{IEEE Transactions on Circuits and Systems for Video Technology}, 2024.

\bibitem{evorl}
J.~Zhang, X.~Yang, W.~He, J.~Ren, Q.~Zhang, Y.~Zhao, R.~Bai, X.~He, and J.~Liu, ``Scale optimization using evolutionary reinforcement learning for object detection on drone imagery,'' in \emph{Proceedings of the AAAI Conference on Artificial Intelligence}, vol.~38, no.~1, 2024, pp. 410--418.

\bibitem{aerial-tcsvt2}
F.~Lin, C.~Fu, Y.~He, F.~Guo, and Q.~Tang, ``Learning temporary block-based bidirectional incongruity-aware correlation filters for efficient uav object tracking,'' \emph{IEEE Transactions on Circuits and Systems for Video Technology}, vol.~31, no.~6, pp. 2160--2174, 2020.

\bibitem{aerial-tcsvt4}
Y.~Yao, G.~Cheng, C.~Lang, X.~Xie, and J.~Han, ``Centric probability-based sample selection for oriented object detection,'' \emph{IEEE Transactions on Circuits and Systems for Video Technology}, 2024.

\bibitem{dehazing}
W.~Fang, G.~Zhang, Y.~Zheng, and Y.~Chen, ``Multi-task learning for uav aerial object detection in foggy weather condition,'' \emph{Remote Sensing}, vol.~15, no.~18, p. 4617, 2023.

\bibitem{derain}
Y.~Xi, W.~Jia, Q.~Miao, J.~Feng, X.~Liu, and F.~Li, ``Coderainnet: Collaborative deraining network for drone-view object detection in rainy weather conditions,'' \emph{Remote Sensing}, vol.~15, no.~6, p. 1487, 2023.

\bibitem{humanbrain}
C.~G. Bartnik and I.~I. Groen, ``Visual perception in the human brain: How the brain perceives and understands real-world scenes,'' in \emph{Oxford Research Encyclopedia of Neuroscience}, 2023.

\bibitem{tph}
X.~Zhu, S.~Lyu, X.~Wang, and Q.~Zhao, ``Tph-yolov5: Improved yolov5 based on transformer prediction head for object detection on drone-captured scenarios,'' in \emph{Proceedings of the IEEE/CVF International Conference on Computer Vision}, 2021, pp. 2778--2788.

\bibitem{aerial-tcsvt5}
K.~Cheng, H.~Cui, H.~A. Ghafoor, H.~Wan, Q.~Mao, and Y.~Zhan, ``Tiny object detection via regional cross self-attention network,'' \emph{IEEE Transactions on Circuits and Systems for Video Technology}, 2022.

\bibitem{ppnet}
W.~Wu, H.~Chang, Z.~Chen, and Z.~Li, ``Plug-and-play robust aerial object detection under hazy conditions,'' \emph{IEEE Journal of Selected Topics in Applied Earth Observations and Remote Sensing}, 2024.

\bibitem{detclip}
L.~Yao, J.~Han, Y.~Wen, X.~Liang, D.~Xu, W.~Zhang, Z.~Li, C.~Xu, and H.~Xu, ``Detclip: Dictionary-enriched visual-concept paralleled pre-training for open-world detection,'' \emph{Advances in Neural Information Processing Systems}, vol.~35, pp. 9125--9138, 2022.

\bibitem{tcsvt-lang}
R.~Liang, Y.~Li, J.~Zhou, and X.~Li, ``Text-driven traffic anomaly detection with temporal high-frequency modeling in driving videos,'' \emph{IEEE Transactions on Circuits and Systems for Video Technology}, 2024.

\bibitem{da-pro}
H.~Li, R.~Zhang, H.~Yao, X.~Song, Y.~Hao, Y.~Zhao, L.~Li, and Y.~Chen, ``Learning domain-aware detection head with prompt tuning,'' \emph{Advances in Neural Information Processing Systems}, vol.~36, 2024.

\bibitem{tcsvt-lang2}
S.~Xu, X.~Li, S.~Wu, W.~Zhang, Y.~Tong, and C.~C. Loy, ``Dst-det: Open-vocabulary object detection via dynamic self-training,'' \emph{IEEE Transactions on Circuits and Systems for Video Technology}, 2024.

\bibitem{llms-vlms}
S.~Jin, X.~Jiang, J.~Huang, L.~Lu, and S.~Lu, ``Llms meet vlms: Boost open vocabulary object detection with fine-grained descriptors,'' in \emph{The Twelfth International Conference on Learning Representations}, 2023.

\bibitem{generateu}
L.~Chuang, J.~Yi, Q.~Lizhen, Y.~Zehuan, and C.~Jianfei, ``Generative region-language pretraining for open-ended object detection,'' in \emph{Proceedings of the IEEE/CVF Conference on Computer Vision and Pattern Recognition}, 2024.

\bibitem{lenna}
F.~Wei, X.~Zhang, A.~Zhang, B.~Zhang, and X.~Chu, ``Lenna: Language enhanced reasoning detection assistant,'' \emph{arXiv preprint arXiv:2312.02433}, 2023.

\bibitem{detgpt}
R.~Pi, J.~Gao, S.~Diao, R.~Pan, H.~Dong, J.~Zhang, L.~Yao, J.~Han, H.~Xu, L.~Kong \emph{et~al.}, ``Detgpt: Detect what you need via reasoning,'' in \emph{The 2023 Conference on Empirical Methods in Natural Language Processing}, 2023.

\bibitem{blip2}
J.~Li, D.~Li, S.~Savarese, and S.~Hoi, ``Blip-2: Bootstrapping language-image pre-training with frozen image encoders and large language models,'' in \emph{International Conference on Machine Learning}.\hskip 1em plus 0.5em minus 0.4em\relax PMLR, 2023, pp. 19\,730--19\,742.

\bibitem{vicuna}
W.-L. Chiang, Z.~Li, Z.~Lin, Y.~Sheng, Z.~Wu, H.~Zhang, L.~Zheng, S.~Zhuang, Y.~Zhuang, J.~E. Gonzalez \emph{et~al.}, ``Vicuna: An open-source chatbot impressing gpt-4 with 90\%* chatgpt quality,'' \emph{See https://vicuna. lmsys. org (accessed 14 April 2023)}, vol.~2, no.~3, p.~6, 2023.

\bibitem{grounding-dino}
S.~Liu, Z.~Zeng, T.~Ren, F.~Li, H.~Zhang, J.~Yang, C.~Li, J.~Yang, H.~Su, J.~Zhu \emph{et~al.}, ``Grounding dino: Marrying dino with grounded pre-training for open-set object detection,'' \emph{arXiv preprint arXiv:2303.05499}, 2023.

\bibitem{clipthegap}
V.~Vidit, M.~Engilberge, and M.~Salzmann, ``Clip the gap: A single domain generalization approach for object detection,'' in \emph{Proceedings of the IEEE/CVF Conference on Computer Vision and Pattern Recognition}, 2023, pp. 3219--3229.

\bibitem{uavdt}
D.~Du, Y.~Qi, H.~Yu, Y.~Yang, K.~Duan, G.~Li, W.~Zhang, Q.~Huang, and Q.~Tian, ``The unmanned aerial vehicle benchmark: Object detection and tracking,'' in \emph{Proceedings of the European Conference on Computer Vision}, 2018, pp. 370--386.

\bibitem{visdrone}
P.~Zhu, L.~Wen, D.~Du, X.~Bian, H.~Fan, Q.~Hu, and H.~Ling, ``Detection and tracking meet drones challenge,'' \emph{IEEE Transactions on Pattern Analysis and Machine Intelligence}, vol.~44, no.~11, pp. 7380--7399, 2021.

\bibitem{delving}
Z.~Wu, K.~Suresh, P.~Narayanan, H.~Xu, H.~Kwon, and Z.~Wang, ``Delving into robust object detection from unmanned aerial vehicles: A deep nuisance disentanglement approach,'' in \emph{Proceedings of the IEEE/CVF International Conference on Computer Vision}, 2019, pp. 1201--1210.

\bibitem{meteor}
B.-K. Lee, C.~W. Kim, B.~Park, and Y.~M. Ro, ``Meteor: Mamba-based traversal of rationale for large language and vision models,'' \emph{arXiv preprint arXiv:2405.15574}, 2024.

\bibitem{coco}
T.-Y. Lin, M.~Maire, S.~Belongie, J.~Hays, P.~Perona, D.~Ramanan, P.~Doll{\'a}r, and C.~L. Zitnick, ``Microsoft coco: Common objects in context,'' in \emph{Computer Vision--ECCV 2014: 13th European Conference, Zurich, Switzerland, September 6-12, 2014, Proceedings, Part V 13}.\hskip 1em plus 0.5em minus 0.4em\relax Springer, 2014, pp. 740--755.

\bibitem{pytorch}
A.~Paszke, S.~Gross, F.~Massa, A.~Lerer, J.~Bradbury, G.~Chanan, T.~Killeen, Z.~Lin, N.~Gimelshein, L.~Antiga \emph{et~al.}, ``Pytorch: An imperative style, high-performance deep learning library,'' \emph{Advances in Neural Information Processing Systems}, vol.~32, 2019.

\bibitem{rtdetr}
Y.~Zhao, W.~Lv, S.~Xu, J.~Wei, G.~Wang, Q.~Dang, Y.~Liu, and J.~Chen, ``Detrs beat yolos on real-time object detection,'' 2023.

\bibitem{ddq}
S.~Zhang, X.~Wang, J.~Wang, J.~Pang, C.~Lyu, W.~Zhang, P.~Luo, and K.~Chen, ``Dense distinct query for end-to-end object detection,'' in \emph{Proceedings of the IEEE/CVF Conference on Computer Vision and Pattern Recognition}, 2023, pp. 7329--7338.

\bibitem{mistral-7b}
A.~Q. Jiang, A.~Sablayrolles, A.~Mensch, C.~Bamford, D.~S. Chaplot, D.~d.~l. Casas, F.~Bressand, G.~Lengyel, G.~Lample, L.~Saulnier \emph{et~al.}, ``Mistral 7b,'' \emph{arXiv preprint arXiv:2310.06825}, 2023.

\bibitem{mist-7b-inst-v0.3}
A.~Jiang, A.~Sablayrolles, A.~Tacnet, A.~Roux, A.~Mensch, A.~Herblin-Stoop, B.~Bout, B.~de~Monicault, B.~Savary, Bam4d, C.~Feldman, D.~S. Chaplot, D.~de~las Casas, E.~Arcelin, E.~B. Hanna, E.~Metzger, G.~Lengyel, G.~Bour, G.~Lample, H.~Rajaona, J.-M. Delignon, J.~Li, J.~Murke, L.~Martin, L.~Ternon, L.~Saulnier, L.~R. Lavaud, M.~Jennings, M.~Pellat, M.~Torelli, M.-A. Lachaux, N.~Schuhl, P.~von Platen, P.~Stock, S.~Subramanian, S.~Yang, S.~Antoniak, T.~L. Scao, T.~Lavril, T.~Lacroix, T.~Gervet, T.~Wang, V.~Nemychnikova, W.~E. Sayed, and W.~Marshall, ``Mistral-7b-instruct-v0.3,'' 2024, \url{https://huggingface.co/mistralai/Mistral-7B-Instruct-v0.3} \textit{(accessed 15 June 2024)}.

\bibitem{mpnet}
K.~Song, X.~Tan, T.~Qin, J.~Lu, and T.-Y. Liu, ``Mpnet: Masked and permuted pre-training for language understanding,'' \emph{Advances in Neural Information Processing Systems}, vol.~33, pp. 16\,857--16\,867, 2020, \url{https://huggingface.co/sentence-transformers/all-mpnet-base-v2} \textit{(accessed 15 June 2024)}.

\bibitem{cadnet}
K.~Duan, D.~Du, D.~H. Qi, and Q.~Huang, ``Detecting small objects using a channel-aware deconvolutional network,'' \emph{IEEE Transactions on Circuits and Systems for Video Technology}, vol.~30, no.~6, pp. 1639--1652, 2019.

\bibitem{ganet}
C.~YuanQiang, D.~Du, L.~Zhang, L.~Wen, W.~Wang, Y.~Wu, and S.~Lyu, ``Guided attention network for object detection and counting on drones,'' in \emph{Proceedings of the 28th ACM International Conference on Multimedia}, 2020, pp. 709--717.

\bibitem{osr-fpn}
X.~Tang, Q.~Yang, D.~Xiong, Y.~Xie, H.~Wang, and R.~Li, ``Improving multiscale object detection with off-centered semantics refinement,'' \emph{IEEE Transactions on Circuits and Systems for Video Technology}, vol.~32, no.~10, pp. 6888--6899, 2022.

\bibitem{a-ndft}
C.~Lee, J.~Seo, and H.~Jung, ``Training domain-invariant object detector faster with feature replay and slow learner,'' in \emph{Proceedings of the IEEE/CVF Conference on Computer Vision and Pattern Recognition}, 2021, pp. 1172--1181.

\bibitem{spotnet}
H.~Perreault, G.-A. Bilodeau, N.~Saunier, and M.~H{\'e}ritier, ``Spotnet: Self-attention multi-task network for object detection,'' in \emph{2020 17th Conference on Computer and Robot Vision}.\hskip 1em plus 0.5em minus 0.4em\relax IEEE, 2020, pp. 230--237.

\bibitem{flsl}
Q.~Su, A.~Netchaev, H.~Li, and S.~Ji, ``Flsl: Feature-level self-supervised learning,'' \emph{Advances in Neural Information Processing Systems}, vol.~36, 2023.

\bibitem{focus}
O.~C. Koyun, R.~K. Keser, I.~B. Akkaya, and B.~U. T{\"o}reyin, ``Focus-and-detect: A small object detection framework for aerial images,'' \emph{Signal Processing: Image Communication}, vol. 104, p. 116675, 2022.

\bibitem{dshnet}
W.~Yu, T.~Yang, and C.~Chen, ``Towards resolving the challenge of long-tail distribution in uav images for object detection,'' in \emph{Proceedings of the IEEE/CVF Winter Conference on Applications of Computer Vision}, 2021, pp. 3258--3267.

\bibitem{fldet}
S.~Wang, K.~Liu, J.~Huang, and X.~Li, ``Fldet: Faster and lighter aerial object detector,'' \emph{IEEE Transactions on Circuits and Systems for Video Technology}, 2024.

\bibitem{yolc}
C.~Liu, G.~Gao, Z.~Huang, Z.~Hu, Q.~Liu, and Y.~Wang, ``Yolc: You only look clusters for tiny object detection in aerial images,'' \emph{IEEE Transactions on Intelligent Transportation Systems}, 2024.

\bibitem{ogmn}
X.~Li, W.~Diao, Y.~Mao, P.~Gao, X.~Mao, X.~Li, and X.~Sun, ``Ogmn: Occlusion-guided multi-task network for object detection in uav images,'' \emph{ISPRS Journal of Photogrammetry and Remote Sensing}, vol. 199, pp. 242--257, 2023.

\bibitem{prdet}
J.~Leng, M.~Mo, Y.~Zhou, C.~Gao, W.~Li, and X.~Gao, ``Pareto refocusing for drone-view object detection,'' \emph{IEEE Transactions on Circuits and Systems for Video Technology}, vol.~33, no.~3, pp. 1320--1334, 2022.

\end{thebibliography}

\end{document}